\documentclass[runningheads]{llncs}

\usepackage{tikz}
\usepackage{bm}
\usepackage{comment}
\usepackage{amsmath,amssymb} 
\usepackage{color}
\usepackage{hyperref}

\usepackage{multirow}
\usepackage{graphics}
\usepackage{caption}
\usepackage{graphicx}
\usepackage{booktabs}       
\usepackage{amsfonts}       
\usepackage{nicefrac}       
\usepackage{microtype}      
\usepackage{xcolor}         

\usepackage[accsupp]{axessibility}  
\usepackage{orcidlink}

\newcommand{\red}[1]{{\color{black} #1}}

\begin{document}
\pagestyle{headings}
\mainmatter
\def\ECCVSubNumber{6076}  

\title{Towards Scale-Aware, Robust, and Generalizable Unsupervised Monocular Depth Estimation by Integrating IMU Motion Dynamics} 

\titlerunning{IMU-Integrated Unsupervised Monocular Depth Estimation}
%
\author{Sen Zhang\orcidlink{0000-0002-8065-5095}\and
Jing Zhang\orcidlink{0000-0001-6595-7661}\and
Dacheng Tao\orcidlink{0000-0001-7225-5449}}
\authorrunning{S. Zhang, J. Zhang, and D.Tao}
%
\institute{The University of Sydney, Sydney, Australia \\
\email{szha2609@uni.sydney.edu.au}\\ \email{\{jing.zhang1,dacheng.tao\}@sydney.edu.au}}

\maketitle

\begin{abstract}
Unsupervised monocular depth and ego-motion estimation has drawn extensive research attention in recent years. Although current methods have reached a high up-to-scale accuracy, they usually fail to learn the true scale metric due to the inherent scale ambiguity from training with monocular sequences. In this work, we tackle this problem and propose DynaDepth, a novel scale-aware framework that integrates information from vision and IMU motion dynamics. Specifically, we first propose an IMU photometric loss and a cross-sensor photometric consistency loss to provide dense supervision and absolute scales. To fully exploit the complementary information from both sensors, we further drive a differentiable camera-centric extended Kalman filter (EKF) to update the IMU preintegrated motions when observing visual measurements. In addition, the EKF formulation enables learning an ego-motion uncertainty measure, which is non-trivial for unsupervised methods. By leveraging IMU during training, DynaDepth not only learns an absolute scale, but also provides a better generalization ability and robustness against vision degradation such as illumination change and moving objects. We validate the effectiveness of DynaDepth by conducting extensive experiments and simulations on the KITTI and Make3D datasets. (\href{https://github.com/SenZHANG-GitHub/ekf-imu-depth}{code})

\end{abstract}

\section{Introduction}

Monocular depth estimation is a fundamental computer vision task which plays an essential role in many real-world applications such as autonomous driving, robot navigation, and virtual reality~\cite{taketomi2017visual,khan2020deep,zhang2020empowering}. Classical geometric methods resolve this problem by leveraging the geometric relationship between temporally contiguous frames and formulating depth prediction as an optimization problem~\cite{engel2014lsd,mur2015orb,engel2017direct}. While geometric methods have achieved good performance, they are sensitive to either textureless regions or illumination changes. The computational cost for dense depth prediction also limits their practical use. Recently deep learning techniques have reformed this research field by training networks to predict depth directly from monocular images and designing proper losses based on ground-truth depth labels or geometric depth clues from visual data. While supervised learning methods achieve the best performance~\cite{eigen2014depth,liu2015learning,fu2018deep,bhat2021adabins,zhang2022information}, the labour cost for collecting ground-truth labels prohibits their use in real-world. To address this issue, unsupervised monocular depth estimation has drawn a lot of research attention~\cite{zhou2017unsupervised,godard2019digging}, which leverages the photometric error \red{from backwarping}. 

Although unsupervised monocular depth learning has made great progress in recent years, there still exist several fundamental problems that may obstruct its usage in real-world. First, current methods suffer from the scale ambiguity problem since the backwarping process is equivalent up to an arbitrary scaling factor w.r.t. depth and translation. While current methods are usually evaluated by re-scaling each prediction map using the median ratio between the ground-truth depth and the prediction, it is difficult to obtain such median ratios in practice.  Secondly, it is well-known that the photometric error is sensitive to illumination change and moving objects, which violate the underlying assumption of the backwarping projection. In addition, though uncertainty has been introduced for the photometric error map under the unsupervised learning framework~\cite{klodt2018supervising,yang2020d3vo}, it remains non-trivial to learn an uncertainty measure for the predicted ego-motion, which could further benefit the development of a robust and trustworthy system. 

In this work, we tackle the above-mentioned problems and propose DynaDepth, a novel scale-aware monocular depth and ego-motion prediction method that explicitly integrates IMU \textit{motion dynamics} into the vision-based system under a camera-centric extended Kalman filter (EKF) framework. Modern sensor suites on vehicles that collect data for training neural networks usually contain multiple sensors beyond cameras. IMU presents a commonly-deployed one which is advantageous in that (1) it is robust to the scenarios when vision fails such as in illumination-changing and textureless regions, (2) the absolute scale metric can be recovered by inquiring the IMU motion dynamics, and (3) it does not suffer from the visual domain gap, leading to a better generalization ability across datasets. While integrating IMU information has dramatically improved the performance of classical geometric odometry and simultaneous localization and mapping (SLAM) systems~\cite{mourikis2007multi,leutenegger2015keyframe,qin2018vins}, its potential in the regime of unsupervised monocular depth learning is much less explored, which is the focus of this work.

Specifically, we propose a scale-aware IMU photometric loss which is constructed by performing backwarping using ego-motion integrated from IMU measurements, which provides dense supervision by using the appearance-based photometric loss instead of naively constraining the ego-motion predicted by networks. To accelerate the training process, the IMU preintegration technique~\cite{lupton2011visual,forster2015imu} is adopted to avoid redundant computation. To correct the errors that result from illumination change and moving objects, we further propose a cross-sensor photometric consistency loss between the synthesized target views using network-predicted and IMU-integrated ego-motions, respectively. Unlike classical visual-inertial SLAM systems that accumulate the gravity and the velocity estimates from initial frames, these two metrics are unknown for the image triplet used in unsupervised depth estimation methods. To address this issue, DynaDepth trains two extra lightweight networks that take two consecutive frames as input and predict the camera-centric gravity and velocity during training.

Considering that IMU and camera present two independent sensing modalities that complement each other, we further derive a differentiable camera-centric EKF framework for DynaDepth to fully exploit the potential of both sensors. When observing new ego-motion predictions from visual data, DynaDepth updates the preintegrated IMU terms based on the propagated IMU error states and the covariances of visual predictions. The benefit is two-fold. First, IMU is known to suffer from inherent noises, which could be corrected by the relatively accurate visual predictions. Second, fusing with IMU under the proposed EKF framework not only introduces scale-awareness, but also provides an elegant way to learn an uncertainty measure for the predicted ego-motion, which can be beneficial for recently emerging research methods that incorporate deep learning into classical SLAM systems to achieve the synergy of learning, geometry, and optimization.

\begin{figure}[t]
\centering
\includegraphics[width=\textwidth]{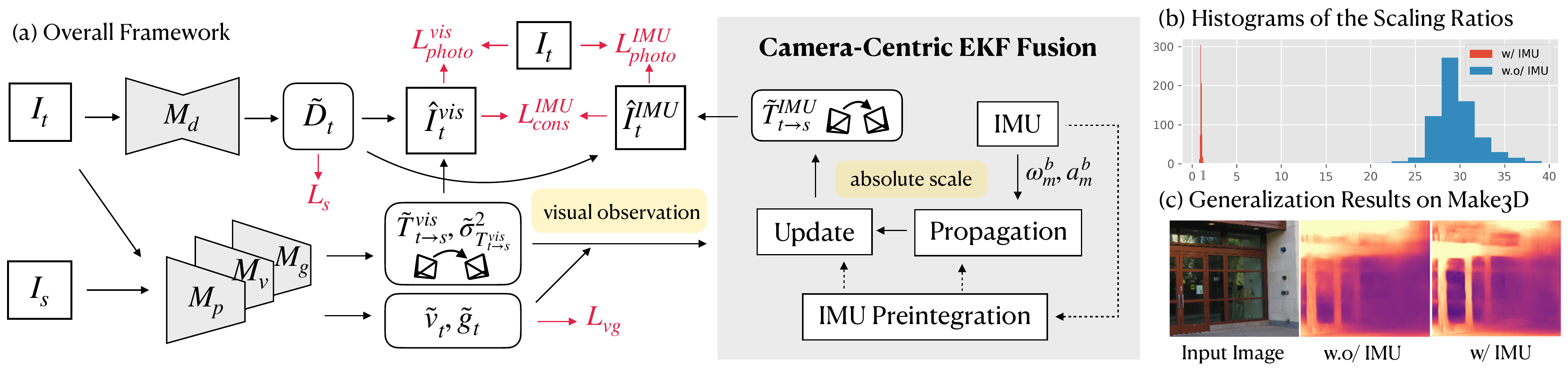}
\caption{(a) The overall framework of DynaDepth. $\hat{I}^{vis}_t$ and $\hat{I}^{IMU}_t$ denote the reconstructed target frames from the source frame $I_s$. Detailed notations of other terms are given in Section~\ref{sec:method}. (b) Histograms of the scaling ratios between the medians of depth predictions and the ground-truth. (c) Generalization results on Make3D using models trained on KITTI with (w/) and without (w.o/) IMU.}
\label{fig:framework}
\end{figure}

Our overall framework is shown in Fig.~\ref{fig:framework}. In summary, our contributions are:
\begin{itemize}
    \item We propose an IMU photometric loss and a cross-sensor photometric consistency loss to provide dense supervision and absolute scales
    \item We derive a differentiable camera-centric EKF framework for sensor fusion.
    \item We show that DynaDepth benefits (1) the learning of the absolute scale, (2) the generalization ability, (3) the robustness against vision degradation such as illumination change and moving objects, and (4) the learning of an ego-motion uncertainty measure, which are also supported by our extensive experiments and simulations on the KITTI and Make3D datasets.
\end{itemize}

\section{Related Work}
\subsection{Unsupervised Monocular Depth Estimation}
Unsupervised monocular depth estimation has drawn extensive research attention recently~\cite{zhou2017unsupervised,mahjourian2018unsupervised,godard2019digging}, which uses the photometric loss by backwarping adjacent images. Recent works improve the performance by introducing multiple tasks~\cite{yin2018geonet,ranjan2019competitive,jung2021fine}, designing more complex networks and losses~\cite{johnston2020self,guizilini20203d,wang2021can,zhou2021r}, and constructing the photometric loss on learnt features~\cite{shu2020feature}. However, monocular methods suffer from the scale ambiguity problem. DynaDepth tackles this problem by integrating IMU dynamics, which not only provides absolute scale, but also achieves state-of-the-art accuracy even if only lightweight networks are adopted.

\subsection{Scale-Aware Depth Learning }
Though supervised depth learning methods\cite{eigen2014depth,fu2018deep,bhat2021adabins} can predict depths with absolute scale, the cost of collecting ground-truth data limits its practical use. To relieve the scale problem, local reprojected depth consistency loss has been proposed to ensure the scale consistency of the predictions~\cite{bian2019unsupervised,zhao2020towards,zhan2020visual}. However, the absolute scale is not guaranteed in these methods. Similar to DynaDepth, there exist methods that resort to other sensors than monocular camera, such as stereo camera that allows a scale-aware left-right consistency loss~\cite{godard2017unsupervised,godard2019digging,zhang2022towards}, and GPS that provides velocities to constrain the ego-motion network~\cite{guizilini20203d,chawla2021multimodal}. In comparison with these methods, using IMU is beneficial in that (1) IMU provides better generalizability since it does suffer from the visual domain gap, and (2) unlike GPS that cannot be used indoors and cameras that fail in texture-less, dynamic and illumination changing scenes, IMU is more robust to the environments.

\subsection{Visual-Inertial SLAM Systems}
The fusion of vision and IMU has achieved great success in classical visual-inertial SLAM systems~\cite{mourikis2007multi,leutenegger2015keyframe,qin2018vins}, yet this topic is much less explored in learning-based depth and ego-motion estimation. Though recently IMU has been introduced into both supervised~\cite{clark2017vinet,chen2019selective} and unsupervised~\cite{han2019deepvio,shamwell2019unsupervised,wei2021unsupervised} odometry learning, most methods extract IMU features implicitly, while we explicitly utilize IMU dynamics to derive explicit supervisory signals. Li et al.~\cite{li2020towards} and Wagstaff et al.~\cite{wagstaff2022efk_vio} similarly use EKF for odometry learning. Ours differs in that we do not require ground-truth information~\cite{li2020towards} or an initialization step~\cite{wagstaff2022efk_vio} to align the velocities and gravities, but learn these quantities using networks. Instead of expressing the error states in the IMU frame, we further derive a camera-centric EKF framework to facilitate the training process. In addition, compared with odometry methods that do not consider the requirements for depth estimation, we specifically design the losses to provide dense depth supervision for monocular depth estimation.

\section{Methodology}
\label{sec:method}
We present the technical details of DynaDepth in this section. We first revisit the preliminaries of IMU motion dynamics. Then we give the details of camera-centric IMU preintegration and the two IMU-related losses, i.e., the scale-aware IMU photometric loss and the cross-sensor photometric consistency loss. Finally, we present the differentiable camera-centric EKF framework which fuses IMU and camera predictions based on their uncertainties and complements the limitations of each other. A discussion on the connection between DynaDepth and classical visual-inertial SLAM algorithms is also given to provide further insights.

\subsection{IMU Motion Dynamics}
Let $\{\bm{w}_m^b,\bm{a}_m^b\}$ and $\{\bm{w}^b,\bm{a}^w\}$ denote the IMU measurements and the underlying vehicle angular and acceleration. The superscript $b$ and $w$ denote the vector is expressed in the body (IMU) frame or the world frame, respectively. Then we have $\bm{w}_m^b = \bm{w}^b + \bm{b}^g + \bm{n}^g$ and $\bm{a}_m^b = \bm{R}_{bw}(\bm{a}^w+\bm{g}^w) + \bm{b}^a + \bm{n}^a$, where $\bm{g}^w$ is the gravity in the world frame and $\bm{R}_{bw}$ is the rotation matrix from the world frame to the body frame~\cite{huang2019visual}. $\{\bm{b}^g,\bm{b}^a\}$ and $\{\bm{n}^g, \bm{n}^a\}$ denote the Gaussian bias and random walk of the gyroscope and the accelerometer, respectively. Let $\{\bm{p}_{wb_t}, \bm{q}_{wb_t}\}$ and $\bm{v}^w_t$ denote the translation and rotation from the body frame to the world frame, and the velocity expressed in the world frame at time $t$, where $\bm{q}_{wb_t}$ denotes the quaternion. The first-order derivatives of $\{\bm{p}, \bm{v}, \bm{q}\}$ read: $\dot{\bm{p}_{wb_t}} = \bm{v}^w_t$, $\dot{\bm{v}^w_t} = \bm{a}^w_t$, and $\dot{\bm{q}_{wb_t}} = \bm{q}_{wb_t} \otimes [0, \frac{1}{2}\bm{w}^{b_t}]^T$, \red{where $\otimes$ denotes the quaternion multiplication}. Then the continuous IMU motion dynamics from time $i$ to $j$ can be derived as:
\begin{align}
    \bm{p}_{wb_j} &= \bm{p}_{wb_i} + \bm{v}_i^w\Delta t + \int\int_{t\in[i,j]} (\bm{R}_{wb_t}\bm{a}^{b_t} - \bm{g}^w)\red{\mathrm{d}}  t^2, \\
    \bm{v}_j^w &= \bm{v}_i^w + \int_{t\in[i,j]}(\bm{R}_{wb_t}\bm{a}^{b_t} - \bm{g}^w)\red{\red{\mathrm{d}} } t, \\
    \bm{q}_{wb_j} &= \int_{t\in[i,j]}\bm{q}_{wb_t}\otimes [0, \frac{1}{2}\bm{w}^{b_t}]^T \red{\mathrm{d}}  t,
\end{align}
where $\Delta t$ is the time gap between $i$ and $j$. For the discrete cases, we use the averages of $\{\bm{w},\bm{a}\}$ within the time interval to approximate the integrals.

\subsection{The DynaDepth Framework}
DynaDepth aims at jointly training a scale-aware depth network $\mathcal{M}_d$ and an ego-motion network $\mathcal{M}_p$ by fusing IMU and camera information. The overall framework is shown in Fig.~\ref{fig:framework}. Given IMU measurements between two consecutive images, we first recover the camera-centric ego-motion $\{\check{\bm{R}_{c_kc_{k+1}}}, \check{\bm{p}_{c_kc_{k+1}}}\}$ with absolute scale using IMU motion dynamics, and train two network modules $\{\mathcal{M}_g,\mathcal{M}_v\}$ to predict the camera-centric gravity and velocity. Then a scare-aware IMU photometric loss and a cross-sensor photometric consistency loss are built based on the ego-motion from IMU. To complement IMU and camera with each other, DynaDepth further integrates a camera-centric EKF module, leading to an updated ego-motion $\{\hat{\bm{R}_{c_kc_{k+1}}}, \hat{\bm{p}_{c_kc_{k+1}}}\}$ for the two IMU-related losses. 

\subsubsection{IMU Preintegration}
IMU usually collects data at a much higher frequency than camera, i.e., between two image frames there exist multiple IMU records. Since the training losses are defined on ego-motions at the camera frequency, naive use of the IMU motion dynamics requires recalculating the integrals at each training step, which could be computationally expensive. IMU preintegration presents a commonly-used technique to avoid the online integral computation~\cite{lupton2011visual,forster2015imu}, which preintegrates the relative pose increment from the IMU records by leveraging the multiplicative property of rotation, i.e., $\bm{q}_{wb_t} = \bm{q}_{wb_i}\otimes \bm{q}_{b_ib_t}$. Then the integration operations can be put into three preintegration terms which only rely on the IMU measurements and can be precomputed beforehand: (1) $\bm{\alpha}_{b_ib_j} = \int\int_{t\in[i,j]} (\bm{R}_{b_ib_t}\bm{a}^{b_t})\red{\mathrm{d}}  t^2$, (2) $\bm{\beta}_{b_ib_j} = \int_{t\in[i,j]} (\bm{R}_{b_ib_t}\bm{a}^{b_t})\red{\mathrm{d}}  t$, and (3) $\bm{q}_{b_ib_j} = \int_{t\in[i,j]} \bm{q}_{b_ib_t}\otimes [0, \frac{1}{2}\bm{w}^{b_t}]^T \red{\mathrm{d}}  t$. Since IMU preintegration is performed in the IMU body frame while the network predicts ego-motions in the camera fame, we thus establish the discrete camera-centric IMU preintegrated ego-motion as:
\begin{align}
    \check{\bm{R}_{c_kc_{k+1}}} &= \bm{R}_{cb}\mathcal{F}^{-1}(\bm{q}_{b_kb_{k+1}})\bm{R}_{bc}, \\
    \check{\bm{p}_{c_kc_{k+1}}} &= \bm{R}_{cb}\bm{\alpha}_{b_kb_{k+1}} + \check{\bm{R}_{c_kc_{k+1}}}\bm{R}_{cb}\bm{p}_{bc} - \bm{R}_{cb}\bm{p}_{bc} +\tilde{\bm{v}^{c_k}}\Delta t_k - \frac{1}{2}\tilde{\bm{g}^{c_k}}\Delta t_k^2,
\end{align}
where $\mathcal{F}$ denotes the transformation from rotation matrix to quaternion. $\{\bm{R}_{cb},\bm{p}_{cb}\}$ and $\{\bm{R}_{bc},\bm{p}_{bc}\}$ are the extrinsics between the IMU and the camera frames. Of note is the estimation of $\tilde{\bm{v}^{c_k}}$ and $\tilde{\bm{g}^{c_k}}$, which are the velocity and the gravity vectors expressed in the camera frame at time k. 

Classical visual-inertial SLAM systems jointly optimize the velocity and the gravity vectors, and accumulate their estimates from previous steps. A complicated initialization step is usually required to achieve good performance. For unsupervised learning where the training units are randomly sampled short-range clips, it is difficult to apply the aforementioned initialization and accumulation. To address this issue, we propose to predict these two quantities directly from images as well during training, using two extra network modules $\{\mathcal{M}_v,\mathcal{M}_g\}$.

\subsubsection{IMU Photometric Loss}
State-of-the-art visual-inertial SLAM systems usually utilize IMU preintegrated ego-motions by constructing the residues between the IMU preintegrated terms and the system estimates to be optimized. However, naively formulating the training loss as these residues on IMU preintegration terms can only provide sparse supervision for the ego-motion network and thus is inefficient in terms of the entire unsupervised learning system. In this work, we propose an IMU photometric loss $L_{photo}^{IMU}$ to tackle this problem which provides dense supervisory signals for both the depth and the ego-motion networks. Given an image $\bm{I}$ and its consecutive neighbours $\{\bm{I}_{-1},\bm{I}_{1}\}$, $L_{photo}^{IMU}$ reads:
\begin{align}
\small
    L_{photo}^{IMU} &= \frac{1}{N}\sum_{i=1}^N \min_{\delta\in\{-1,1\}} \mathcal{L}(\bm{I}(\bm{y}_i),\bm{I}_{\delta}(\psi(\bm{K}\hat{\bm{R}}_\delta \bm{K}^{-1}\bm{y}_i+\frac{\bm{K}\hat{\bm{p}}_\delta}{\tilde{z_i}}))), \\
    \mathcal{L}(\bm{I},\bm{I}_\delta) &= \alpha\frac{1-SSIM(\bm{I},\bm{I}_\delta)}{2} + (1-\alpha)||\bm{I}-\bm{I}_\delta||_1,
\end{align}
where $\bm{K}$ and $N$ are the camera intrinsics and the number of utilized pixels, $\bm{y}_i$ and $\tilde{z_i}$ are the pixel coordinate in image $\bm{I}$ and its depth predicted by $\mathcal{M}_d$, \red{$\bm{I}(\bm{y}_i)$ is the pixel intensity at $\bm{y}_i$}, and $\psi(\cdot)$ denotes the depth normalization function. $\{\hat{\bm{R}}_\delta,\hat{\bm{p}}_\delta\}$ denotes the ego-motion estimate from image $\bm{I}$ to $\bm{I}_\delta$, which is obtained by fusing the IMU preintegrated ego-motion and the ones predicted by $\mathcal{M}_p$ under our camera-centric EKF framework. $SSIM(\cdot)$ denotes the structural similarity index~\cite{wang2004image}. We also adopt the per-pixel minimum trick proposed in \cite{godard2019digging}.

\subsubsection{Cross-Sensor Photometric Consistency Loss}
In addition to $L_{photo}^{IMU}$, we further propose a cross-sensor photometric consistency loss $L_{photo}^{cons}$ to align the ego-motions from IMU preintegration and $\mathcal{M}_p$. Instead of directly comparing the ego-motions, we use the photometric error between the backwarped images, which provides denser supervisory signals for both $\mathcal{M}_d$ and $\mathcal{M}_p$:
\begin{equation}
\small
    L_{photo}^{cons} = \frac{1}{N}\sum_{i=1}^N \min_{\delta\in\{-1,1\}} \mathcal{L}(\bm{I}_{\delta}(\psi(\bm{K}\tilde{\bm{R}}_\delta \bm{K}^{-1}\bm{y}_i+\frac{\bm{K}\tilde{\bm{p}}_\delta}{\tilde{z_i}})),\bm{I}_{\delta}(\psi(\bm{K}\hat{\bm{R}}_\delta \bm{K}^{-1}\bm{y}_i+\frac{\bm{K}\hat{\bm{p}}_\delta}{\tilde{z_i}}))),
\end{equation}
where $\{\tilde{\bm{R}}_\delta,\tilde{\bm{p}}_\delta\}$ are the ego-motion predicted by $\mathcal{M}_p$. 
\paragraph{Remark:} Of note is that using $L_{photo}^{cons}$  actually increases the tolerance for illumination change and moving objects which may violate the underlying assumption of the photometric loss between consecutive frames. Since we are comparing two backwarped views in $L_{photo}^{cons}$, the errors incurred by the corner cases will be exhibited equally in both backwarped views. In this sense, $L_{photo}^{cons}$ remains valid, and minimizing $L_{photo}^{cons}$ helps to align $\{\tilde{\bm{R}}_\delta,\tilde{\bm{p}}_\delta\}$ and $\{\hat{\bm{R}}_\delta,\hat{\bm{p}}_\delta\}$ under such cases.

\subsubsection{The Camera-Centric EKF Fusion}
To fully exploit the complementary IMU and camera sensors, we propose to fuse ego-motions from both sensors under a camera-centric EKF framework. Different from previous methods that integrate EKF into deep learning-based frameworks to deal with IMU data~\cite{liu2020tlio,li2020towards}, ours differs in that we do not require ground-truth ego-motion and velocities to obtain the aligned velocities and gravities for each IMU frame, but propose $\{\mathcal{M}_v,\mathcal{M}_g\}$ to predict these quantities. In addition, instead of expressing the error states in the IMU body frame, we derive the camera-centric EKF propagation and update processes to facilitate the training process \red{which takes camera images as input}.

\paragraph{EKF Propagation:} Let $c_k$ denote the camera frame at time $t_k$, and $\{b_t\}$ denote the IMU frames between $t_k$ and time $t_{k+1}$ when we receive the next visual measurement. We then propagate  the IMU information according to the state transition model: $\bm{x}_t = f(\bm{x}_{t-1}, \bm{u}_t) + \bm{w}_t$, where $\bm{u}_t$ is the IMU record at time $t$, $\bm{w}_t$ is the noise term, and $\bm{x}_t=[\bm{\phi}_{c_kb_t}^T, \bm{p}_{c_kb_t}^T, \bm{v}^{c_kT}, \bm{g}^{c_kT}, \bm{b}_w^{b_tT}, \bm{b}_a^{b_tT}]^T$ is the state vector expressed in the camera frame $c_k$ except for $\{\bm{b}_w,\bm{b}_a\}$. $\bm{\phi}_{c_kb_t}$ denotes the so(3) Lie algebra of the rotation matrix $\bm{R}_{c_kb_t}$ s.t. $\bm{R}_{c_kb_t} = exp([\bm{\phi}_{c_kb_t}]^\wedge)$, where $[\cdot]^\wedge$ denotes the operation from a so(3) vector to the corresponding skew symmetric matrix. To facilitate the derivation of the propagation process, we further separate the state into the nominal states denoted by $\bar{(\cdot)}$, and the error states $\delta \bm{x}_{b_t}=[\delta \bm{\phi}_{c_kb_t}^T, \delta \bm{p}_{c_kb_t}^T, \delta \bm{v}^{c_kT}, \delta \bm{g}^{c_kT}, \delta \bm{b}_w^{b_tT}, \delta \bm{b}\bm{_}a^{b_tT}]^T$, such that:
\begin{align}
    \bm{R}_{c_kb_t} &= \bar{\bm{R}}_{c_kb_t}exp([\delta \bm{\phi}_{c_kb_t}]^\wedge), \quad \bm{p}_{c_kb_t} = \bar{\bm{p}}_{c_kb_t} + \delta \bm{p}_{c_kb_t}, \label{eq:error_0} \\
    \bm{v}^{c_k} &= \bar{\bm{v}}^{c_k} + \delta \bm{v}^{c_k}, \quad \bm{g}^{c_k} = \bar{\bm{g}}^{c_k} + \delta \bm{g}^{c_k}, \\
    \bm{b}_w^{b_t} &= \bar{\bm{b}_w}^{b_t} + \delta \bm{b}_w^{b_t}, \quad \bm{b}_a^{b_t} = \bar{\bm{b}_a}^{b_t} + \delta \bm{b}_a^{b_t} \label{eq:error_1}.
\end{align}

The nominal states can be computed using the preintegration terms, while the error states are used for propagating the covariances. It is noteworthy that the state transition model of $\delta \bm{x}_{b_t}$ is non-linear, which prevents a naive use of the Kalman filter. EKF addresses this problem and performs propagation by linearizing the state transition model at each time step using the first-order Taylor approximation. Therefore, let $\dot{(\cdot)}$ denote the derivative w.r.t. time $t$, we derive the continuous-time propagation model for the error states as: $\delta \dot{\bm{x}}_{b_t} = \bm{F} \delta \bm{x}_{b_t} + \bm{G} \bm{n}$. Detailed derivations are given in the Supplementary material, and $\bm{F}$ and $\bm{G}$ read:
\begin{equation}
\small
    \bm{F} = \begin{bmatrix}
            -[\bar{\bm{w}}^{b_t}]^\wedge & 0 & 0 & 0 & -\bm{I}_3 & 0 \\
            0 & 0 & \bm{I}_3 & 0 & 0 & 0 \\
            -\bar{\bm{R}}_{c_kb_t}[\bar{\bm{R}}_{c_kb_t}^T\bar{\bm{g}}^{c_k} + \bar{\bm{a}}^{b_t}]^\wedge & 0 & 0 & -\bm{I}_3 & 0 & -\bar{\bm{R}}_{c_kb_t} \\
            0 & 0 & 0 & 0 & 0 & 0 \\
            0 & 0 & 0 & 0 & 0 & 0 \\
            0 & 0 & 0 & 0 & 0 & 0 
        \end{bmatrix},
        \bm{G} = \begin{bmatrix}
        -\bm{I}_3 & 0 & 0 & 0 \\
        0 & 0 & 0 & 0 \\
        0 & 0 & -\bar{\bm{R}}_{c_kb_t} & 0 \\
        0 & 0 & 0 & 0 \\
        0 & \bm{I}_3 & 0 & 0 \\
        0 & 0 & 0 & \bm{I}_3 
    \end{bmatrix}
\end{equation}
where $\bar{\bm{w}}^{b_t} = \bm{w}_m^{b_t} - \bar{\bm{b}_w}^{b_t}$ and $\bar{\bm{a}}^{b_t} = \bm{a}_m^{b_t} - \bar{\bm{R}}_{c_kb_t}^T\bar{\bm{g}}_{c_k} - \bar{\bm{b}_a}^{b_t}$. Given the continuous error propagation model and the initial condition $\bm{\Phi}_{t_\tau, t_\tau}=\bm{I}_{18}$, the discrete state-transition matrix $\bm{\Phi}_{(t_{\tau+1},t_\tau)}$ can be found by solving $\dot{\bm{\Phi}}_{(t_{\tau+1}, t_\tau)} = \bm{F}_{t_{\tau+1}} \bm{\Phi}_{(t_{\tau+1}, t_\tau)}$:
\begin{equation}
    \bm{\Phi}_{t_{\tau+1}, t_\tau} = exp(\int_{t_\tau}^{t_{\tau+1}} \bm{F}(s) \red{\mathrm{d}}  s) \approx \bm{I}_{18} + \bm{F}\delta t + \frac{1}{2} \bm{F}^2\delta t^2, \ \ \ \delta t = t_{\tau+1} - t_\tau.
\end{equation}

Let $\check{\bm{P}}$ and $\hat{\bm{P}}$ denote the prior and posterior covariance estimates during propagation and after an update given new observations. Then we have
\begin{align}
    \check{\bm{P}_{t_{\tau+1}}} &= \bm{\Phi}_{t_{\tau+1}, t_\tau}\check{\bm{P}_{t_\tau}}\bm{\Phi}_{t_{\tau+1}, t_\tau}^T + \bm{Q}_{t_\tau}, \\
    \bm{Q}_{t_\tau} &= \int_{t_\tau}^{t_{\tau+1}} \bm{\Phi}_{s,t_\tau}\bm{G}\bm{Q}\bm{G}^T\bm{\Phi}_{s,t_\tau}^T \red{\mathrm{d}}  s \approx \bm{\Phi}_{t_{\tau+1},t_\tau}\bm{G}\bm{Q}\bm{G}^T\bm{\Phi}_{t_{\tau+1},t_\tau}^T \delta t,
\end{align}
where $\bm{Q}=\mathcal{D}([\sigma^2_w \bm{I}_3,\sigma^2_{b_w} \bm{I}_3,\sigma^2_{a} \bm{I}_3, \sigma^2_{b_a} \bm{I}_3 ])$.  $\mathcal{D}$ is the diagonalization function.

\paragraph{EKF Update:} 
In general, given an observation measurement $\bm{\xi}_{k+1}$ and its corresponding covariance $\bm{\Gamma}_{k+1}$ from the camera sensor at time $t_{k+1}$, we assume the following observation model: $\bm{\xi}_{k+1} = h(\bm{x}_{k+1}) + \bm{n}_r,\ \bm{n}_r\sim N(0, \bm{\Gamma}_{k+1})$.

Let $\bm{H}_{k+1} = \frac{\partial h(\bm{x}_{k+1})}{\partial \delta \bm{x}_{k+1}}$. Then the EKF update applies as following:
\begin{align}
    \bm{K}_{k+1} &= \check{\bm{P}}_{k+1} \bm{H}_{k+1}^T(\bm{H}_{k+1}\check{\bm{P}_{k+1}}\bm{H}_{k+1}^T + \bm{\Gamma}_{k+1})^{-1}, \label{eq: ekf_update_0}\\
    \hat{\bm{P}}_{k+1} &= (\bm{I}_{18} - \bm{K}_{k+1}\bm{H}_{k+1})\check{\bm{P}}_{k+1}, \\
    \delta \hat{\bm{x}}_{k+1} &= \bm{K}_{k+1} (\bm{\xi}_{k+1} - h(\check{\bm{x}}_{k+1})). \label{eq: ekf_update_1}
\end{align}

In DynaDepth, the observation measurement is defined as the ego-motion predicted by $\mathcal{M}_p$, i.e., $\bm{\xi}_{k+1} = [\tilde{\bm{\phi}}_{c_kc_{k+1}}^T,\tilde{\bm{p}}_{c_kc_{k+1}}^T]^T$. Of note is that the covariances $\bm{\Gamma}_{k+1}$ of $\{\tilde{\bm{\phi}}_{c_kc_{k+1}}^T,\tilde{\bm{p}}_{c_kc_{k+1}}^T\}$ are also predicted by the ego-motion network $\mathcal{M}_p$. To finish the camera-centric EKF update step, we derive $h(\check{\bm{x}}_{k+1})$ and $\bm{H}_{k+1}$ as:
\begin{equation}
\small
    h(\check{\bm{x}}_{k+1}) = \begin{bmatrix}
        \bar{\bm{\phi}}_{c_kc_{k+1}} \\
        \bar{\bm{R}}_{c_kb_{k+1}}\bm{p}_{bc} + \bar{\bm{p}}_{c_kb_{k+1}}
    \end{bmatrix},
    \bm{H}_{k+1} = \begin{bmatrix}
        J_l(-\bar{\bm{\phi}}_{c_kc_{k+1}})^{-1}\bm{R}_{cb} & 0 & 0 & 0 & 0 & 0 \\
        -\bar{\bm{R}}_{c_kb_{k+1}}[\bm{p}_{bc}]^\wedge & \bm{I}_3 & 0 & 0 & 0 & 0 
    \end{bmatrix}. \label{eq:cam_ekf_update}
\end{equation}
After obtaining the updated error states $\delta \hat{\bm{x}}_{k+1}$, we add $\delta \hat{\bm{x}}_{k+1}$ back to the accumulated nominal states to get the corrected ego-motion. In detail, $\delta \hat{\bm{x}}_{k+1}$ is obtained by inserting Eq. (\ref{eq:cam_ekf_update}) into  Eq. (\ref{eq: ekf_update_0}-\ref{eq: ekf_update_1}), which can be inserted into Eq. (\ref{eq:error_0}) to get the updated $\{\hat{\bm{\phi}}_{c_kb_{k+1}}, \hat{\bm{p}}_{c_kb_{k+1}}\}$. Then by projecting $\{\hat{\bm{\phi}}_{c_kb_{k+1}}, \hat{\bm{p}}_{c_kb_{k+1}}\}$ using the camera intrinsics, we obtain the corrected ego-motion $\{\hat{\bm{\phi}}_{c_kb_{k+1}}, \hat{\bm{p}}_{c_kb_{k+1}}\}$ that fuses IMU and camera information based on their covariances as confidence indicators, which are used to compute $L_{photo}^{IMU}$ and $L_{photo}^{cons}$. 

Finally, in addition to $\{L_{photo}^{IMU},L_{photo}^{cons}\}$, the total training loss $L_{total}$ in DynaDepth also includes the vision-based photometric loss $L_{photo}^{vis}$ and the disparity smoothness loss $L_s$ as proposed in monodepth2~\cite{godard2019digging} to leverage the visual clues. We also consider the weak L2-norm loss $L_{vg}$ for the velocity and gravity predictions from $\mathcal{M}_v$ and $\mathcal{M}_g$. In summary, $L_{total}$ reads:
\begin{equation}
    L_{total} = L_{photo}^{vis} + \lambda_1 L_s + \lambda_2 L_{photo}^{IMU} + \lambda_3 L_{photo}^{cons} + \lambda_4 L_{vg},
\end{equation}
where $\{\lambda_1,\lambda_2,\lambda_3,\lambda_4\}$ denote the loss weights which are determined empirically.

\paragraph{Remark:} Alhough we have witnessed a paradigm shift from EKF to optimization in classical visual-inertial SLAM systems in recent years~\cite{mourikis2007multi,leutenegger2015keyframe,qin2018vins}, we argue that in the setting of unsupervised depth estimation, EKF provides a better choice than optimization. The major problem of EKF is its limited ability to handle long-term data because of the Markov assumption between updates, the first-order approximation for the non-linear state-transition and observation models, and the memory consumption for storing the covariances. However, in our setting, short-term image clips are usually used as the basic training unit, which indicates that the Markov property and the linearization in EKF will approximately hold within the short time intervals. In addition, only the ego-motions predicted by $\mathcal{M}_p$ are used as the visual measurements, which is memory-efficient. 

On the other hand, by using EKF, we are able to correct the IMU preintegrated ego-motions and update $\{L_{photo}^{IMU},L_{photo}^{cons}\}$ accordingly when observing new visual measurements. Compared with formulating the commonly-used optimization objective, i.e., the residues of the IMU preintegration terms, as the training losses, our proposed $L_{photo}^{IMU}$ and $L_{photo}^{cons}$ provide denser supervision for both $\mathcal{M}_d$ and $\mathcal{M}_p$. From another perspective, EKF essentially can be regarded as weighting the ego-motions from IMU and vision based on their covariances, and thus naturally provides a framework for estimating the uncertainty of the ego-motion predicted by $\mathcal{M}_p$, which is non-trivial for the unsupervised learning frameworks.

\section{Experiment}
We evaluate the effectiveness of DynaDepth on KITTI~\cite{geiger2013vision} and test the generalization ability on Make3D~\cite{saxena2008make3d}. In addition, we perform extensive ablation studies on our proposed IMU losses, the EKF framework, the learnt ego-motion uncertainty, and the robustness against illumination change and moving objects.

\subsection{Implementation}
DynaDepth is implemented in pytorch~\cite{paszke2019pytorch}. We adopt the monodepth2~\cite{godard2019digging} network structures for $\{\mathcal{M}_d,\mathcal{M}_p\}$, except that we increase the output dimension of $\mathcal{M}_p$ from 6 to 12 to include the uncertainty predictions. $\{\mathcal{M}_g,\mathcal{M}_v\}$ share the same network structure as $\mathcal{M}_p$ except that the output dimensions are both set to 3. $\{\lambda_1,\lambda_2,\lambda_3,\lambda_4\}$ are set to $\{0.001, 0.5,0.01,0.001\}$. We train all networks for 30 epochs using an initial learning rate 1e-4, which is reduced to 1e-5 after the first 15 epochs. The training process takes $1\sim 2$ days on a single NVIDIA V100 GPU. The source codes and the trained models will be released.

\subsection{Scale-Aware Depth Estimation on KITTI}
We use the Eigen split~\cite{eigen2015predicting} for depth evaluation. In addition to the removal of static frames as proposed in~\cite{zhou2017unsupervised}, we discard images without the corresponding IMU records, leading to 38,102 image-and-IMU triplets for training and 4,238 for validation. WLOG, we use the image resolution 640x192 and cap the depth predictions at 80m, following the common practice in \cite{godard2019digging,johnston2020self,guizilini20203d,chawla2021multimodal,wang2021can}.

\begin{table*}[t]
\caption{Per-image rescaled depth evaluation on KITTI using the Eigen split.  The best and the second best results are shown in \textbf{bold} and \underline{underline}. $^\dagger$ denotes our reproduced results. Results are rescaled using the median ground-truth from Lidar. The means and standard errors of the scaling ratios are reported in Scale.}
\label{tb:main_kitti_1}
\resizebox{\textwidth}{!}{
\begin{tabular}{ccc|cccc|ccc}
\toprule
\multirow{2}*{Methods} & \multirow{2}*{Year} & \multirow{2}*{Scale}  & \multicolumn{4}{c|}{Error$\downarrow$} & \multicolumn{3}{c}{Accuracy$\uparrow$}\\
\cline{4-7}
\cline{8-10}
~ & ~ & ~  & AbsRel & SqRel & RMSE & RMSE$_{log}$ & $\sigma<1.25$ & $\sigma<1.25^2$ & $\sigma<1.25^3$ \\
\midrule
 Monodepth2 R18~\cite{godard2019digging} & ICCV 2019 & NA & 0.112 & 0.851 & 4.754 & 0.190 & 0.881 & 0.960 & \underline{0.981} \\
 Monodepth2 R50$^\dagger$~\cite{godard2019digging} & ICCV 2019 & 29.128$\pm$0.084 & 0.111 & 0.806 & 4.642 & 0.189 & 0.882 & \textbf{0.962} & \textbf{0.982} \\
 PackNet-SfM~\cite{guizilini20203d} & CVPR 2020 & NA & 0.111 & 0.785 & \textbf{4.601} & 0.189 & 0.878 & 0.960 & \textbf{0.982} \\
 Johnston R18~\cite{johnston2020self} & CVPR 2020 & NA & 0.111 & 0.941 & 4.817 & 0.189  & \textbf{0.885} & \underline{0.961} & \underline{0.981}  \\
 R-MSFM6~\cite{zhou2021r}  & ICCV 2021 & NA & 0.112  & 0.806 & 4.704 & 0.191 & 0.878 & 0.960 & \underline{0.981}  \\
 G2S R50~\cite{chawla2021multimodal} & ICRA 2021 & 1.031$\pm$0.073 & 0.112 & 0.894 & 4.852 & 0.192 & 0.877 & 0.958 & \underline{0.981} \\
 ScaleInvariant R18~\cite{wang2021can} & ICCV 2021 & NA & \underline{0.109} & \underline{0.779} & 4.641 & \textbf{0.186} & \underline{0.883}  &  \textbf{0.962}  & \textbf{0.982}  \\
 \midrule 
 DynaDepth R18 & 2022 & \underline{1.021}$\pm$\textbf{0.069} & 0.111 & 0.806 & 4.777 & 0.190 & 0.878  &  0.960  & \textbf{0.982}  \\
 DynaDepth R50 & 2022 & \textbf{1.013}$\pm$\underline{0.071} & \textbf{0.108} & \textbf{0.761} & \underline{4.608} & \underline{0.187} & \underline{0.883}  &  \textbf{0.962}  & \textbf{0.982} \\
\bottomrule
\end{tabular}}
\end{table*}

\begin{table*}[t]
\caption{Unscaled depth evaluation on KITTI using the Eigen split. $^\dagger$ denotes our reproduced results. The best results are shown in \textbf{bold}.}
\label{tb:main_kitti_2}
\resizebox{\textwidth}{!}{
\begin{tabular}{cc|cccc|ccc}
\toprule
\multirow{2}*{Methods} & \multirow{2}*{Year}  & \multicolumn{4}{c|}{Error$\downarrow$} & \multicolumn{3}{c}{Accuracy$\uparrow$}\\
\cline{3-6}
\cline{7-9}
~ & ~ & AbsRel & SqRel & RMSE & RMSE$_{log}$ & $\sigma<1.25$ & $\sigma<1.25^2$ & $\sigma<1.25^3$ \\
\midrule
 Monodepth2 R50$^\dagger$~\cite{godard2019digging} & ICCV 2019  & 0.966 & 15.039 & 19.145 & 3.404 & 0.000 & 0.000 & 0.000 \\
 PackNet-SfM~\cite{guizilini20203d} & CVPR 2020  & 0.111 & 0.829 & 4.788 & 0.199 & 0.864 & 0.954 & 0.980 \\
 G2S R50~\cite{chawla2021multimodal} & ICRA 2021 & \textbf{0.109} & 0.860 & 4.855 & 0.198 & 0.865 & 0.954 & 0.980 \\
 \midrule
 DynaDepth R50 & 2022 &  \textbf{0.109} & \textbf{0.787} & \textbf{4.705} & \textbf{0.195} & \textbf{0.869}  &  \textbf{0.958}  & \textbf{0.981}  \\
\bottomrule
\end{tabular}}
\end{table*}

We compare DynaDepth with state-of-the-art monocular depth estimation methods in Table~\ref{tb:main_kitti_1}, which rescale the results using the ratio of the median depth between the ground-truth and the prediction. For a fair comparison, we only present results achieved with image resolution 640x192 and an encoder with moderate size, i.e., ResNet18 (R18) or ResNet50 (R50). In addition to standard depth evaluation metrics~\cite{eigen2014depth}, we report the means and standard errors of the rescaling factors to demonstrate the scale-awareness ability. DynaDepth achieves the best up-to-scale performance w.r.t. four metrics and achieves the second best for the other three metrics. Of note is that DynaDepth also achieves a nearly perfect absolute scale. In terms of scale-awareness, even our R18 version outperforms G2S R50~\cite{chawla2021multimodal}, which uses a heavier encoder. For better illustration, we also show the scaling ratio histograms with and without IMU in Fig.~\ref{fig:framework}(b). 

We then report the unscaled results in Table~\ref{tb:main_kitti_2}, and compare with PackNet-SfM~\cite{guizilini20203d} and G2S~\cite{chawla2021multimodal}, which use the GPS information to construct velocity constraints. Without rescaling, Monodepth2~\cite{godard2019digging} fails completely as expected. In this case, DynaDepth achieves the best performance w.r.t. all metrics, setting a new benchmark of unscaled depth evaluation for monocular methods.

\subsection{Generalizability on Make3D}
\label{sec:generalizability}
We further test the generalizability of DynaDepth on Make3D~\cite{saxena2008make3d} using models trained on KITTI~\cite{geiger2013vision}. The test images are centre-cropped to a 2x1 ratio for a fair comparison with previous methods~\cite{godard2019digging}. A qualitative example is given in Fig.~\ref{fig:framework}(c), where the model without IMU fails in the glass and shadow areas, while our model achieves a distinguishable prediction. Quantitative results are reported in Table~\ref{tb:main_make3d}. A reasonably good scaling ratio has been achieved for DynaDepth, indicating that the scale-awareness learnt by DynaDepth can be well generalized to unseen datasets. Surprisingly, we found that DynaDepth that only uses the gyroscope and accelerator IMU information (w.o/ $L_{vg}$) achieves the best generalization results. The reason can be two-fold. First, our full model may overfit to the KITTI dataset due to the increased modeling capacity. Second, the performance degradation can be due to the domain gap of the visual data, since both $\mathcal{M}_v$ and $\mathcal{M}_g$ take images as input. This also explains the scale loss of G2S in this case. We further show that DynaDepth w.o/ $L_{vg}$ significantly outperforms the stereo version of Monodepth2, which can also be explained by the visual domain gap, especially the different camera intrinsics used in their left-right consistency loss. Our generalizability experiment justifies the advantages of using IMU to provide scale information, which will not be affected by the visual domain gap and varied camera parameters, leading to improved generalization performance. In addition, it is also shown that the use of EKF in training significantly improves the generalization ability, possibly thanks to the EKF fusion framework that takes the uncertainty into account and integrates the generalizable IMU motion dynamics and the domain-specific vision information in a more reasonable way. 

\begin{table*}[t]
\caption{Generalization results on Make3D. $^*$ denotes unscaled results while the others present per-image rescaled results. The best results are shown in \textbf{bold}. M, S, GPS, and IMU in Type denote whether monocular, stereo, GPS and IMU information are used for training the model on KITTI. - means item not available.} 
\label{tb:main_make3d}
\centering
\resizebox{\textwidth}{!}{
\begin{tabular}{c|cc|cc|cccc|ccc}
\toprule
\multirow{2}*{Methods} & \multirow{2}*{$L_{vg}$} & \multirow{2}*{EKF} & \multirow{2}*{Type} & \multirow{2}*{Scale}  & \multicolumn{4}{c|}{Error$\downarrow$} & \multicolumn{3}{c}{Accuracy$\uparrow$}\\
\cline{6-9}
\cline{10-12}
~ & ~ & ~ &~ & ~ & Abs$_{rel}$ & Sq$_{rel}$ & RMSE & RMSE$_{log}$ & $\sigma<1.25$ & $\sigma<1.25^2$ & $\sigma<1.25^3$ \\
\midrule
 Zhou~\cite{zhou2017unsupervised} & - & - & M & - & 0.383 & 5.321 & 10.470 & 0.478 & - & - & -\\
 Monodepth2~\cite{godard2019digging} & - & - & M & - & 0.322 & 3.589 & 7.417 & 0.163 & - & - & - \\
 G2S~\cite{chawla2021multimodal} & -& - & M+GPS & 2.81$\pm$0.85 & - & - & - & - & - & - & - \\
 DynaDepth &  &  & M+IMU & 1.37$\pm$0.27 & 0.316 & 3.006 & 7.218 & 0.164 & 0.522 & 0.797 & 0.914 \\
 DynaDepth &  & \checkmark & M+IMU & \textbf{1.26}$\pm$0.27 & \textbf{0.313} & \textbf{2.878} & \textbf{7.133} & \textbf{0.162} & \textbf{0.527} & \textbf{0.800} & \textbf{0.916} \\
 DynaDepth (full) & \checkmark & \checkmark  & M+IMU & 1.45$\pm$\textbf{0.26} & 0.334 & 3.311 & 7.463 & 0.169 & 0.497 & 0.779 & 0.908 \\
\midrule 
 Monodepth2$^*$~\cite{godard2019digging} &- &- & M+S & - & 0.374 & 3.792 & 8.238 & \textbf{0.201} & - & - & -\\
 DynaDepth$^*$ &  &  & M+IMU & - & 0.360 & 3.461 & 8.833 & 0.226 & 0.295 & 0.594 & 0.794 \\
 DynaDepth$^*$ &  & \checkmark  & M+IMU & - & \textbf{0.337} & \textbf{3.135} & \textbf{8.217} & \textbf{0.201} & \textbf{0.384} & \textbf{0.671} & \textbf{0.845} \\
 DynaDepth$^*$ (full) & \checkmark & \checkmark & M+IMU & - & 0.378 & 3.655 & 9.034 & 0.240 & 0.261 & 0.550 & 0.758 \\
\bottomrule
\end{tabular}}
\end{table*}

\subsection{Ablation Studies} 

We conduct ablation studies on KITTI to investigate the effects of the proposed IMU-related losses, the EKF fusion framework, and the learnt ego-motion uncertainty. In addition, we design simulated experiment to demonstrate the robustness of DynaDepth against vision degradation such as illumination change and moving objects. WLOG, we use ResNet18 as the encoder for all ablation studies.

\subsubsection{The effects of the IMU-related losses and the EKF Fusion Framework}
We report the ablation results of the IMU-related losses and the EKF fusion framework in Table~\ref{tb:abl_imu_1}. First, $L^{IMU}_{photo}$ presents the main contributor to learning the scale. However, only a rough scale is learnt using $L^{IMU}_{photo}$ only. And the up-to-scale accuracy is also not as good as the other models. $L^{cons}_{photo}$ provides better up-to-scale accuracy, but using $L^{cons}_{photo}$ alone is not enough to learn the absolute scale due to the relatively weak supervision. Instead, combining $L^{IMU}_{photo}$ and $L^{cons}_{photo}$ together boosts the performance of both the scale-awareness and the accuracy. The use of $L_{vg}$ further enhances the evaluation results. Nevertheless, as shown in Section~\ref{sec:generalizability}, $L_{vg}$ may lead to overfitting to current dataset and harm the generalizability, due to its dependence on visual data that suffers from the visual domain gap between different datasets. On the other hand, EKF improves the up-to-scale accuracy w.r.t. almost all metrics, while decreasing the learnt scale information a little bit. Since the scale information comes from IMU, and the visual data contributes most to the up-to-scale accuracy, EKF achieves a good balance between the two sensors. Moreover, as shown in Table~\ref{tb:main_make3d}, the use of EKF leads to the best generalization results w.r.t. both the scale and the accuracy. 

\begin{table*}[t]
\caption{Ablation results of the IMU-related losses and the EKF fusion framework on KITTI. The best results are shown in \textbf{bold}.}
\label{tb:abl_imu_1}
\resizebox{\textwidth}{!}{
\begin{tabular}{cccc|c|cccc|ccc}
\toprule
 \multirow{2}*{EKF} &\multirow{2}*{$L^{IMU}_{photo}$} & \multirow{2}*{$L^{cons}_{photo}$} & \multirow{2}*{$L_{vg}$} & \multirow{2}*{Scale} & \multicolumn{4}{c|}{Error$\downarrow$} & \multicolumn{3}{c}{Accuracy$\uparrow$}\\
\cline{6-9}
\cline{10-12}
 ~ & ~ & ~ & ~ & ~ & AbsRel & SqRel & RMSE & RMSE$_{log}$ & $\sigma<1.25$ & $\sigma<1.25^2$ & $\sigma<1.25^3$ \\
 \midrule
  \checkmark & \checkmark & & & 1.130$\pm$0.099 & 0.115 & 0.804 & 4.806 & 0.193 & 0.871 & 0.959 & \textbf{0.982} \\
  \checkmark & & \checkmark & & 4.271$\pm$0.089 & 0.114 & 0.832 & 4.780 & 0.192 & 0.876 & 0.959 & 0.981 \\
  \checkmark & \checkmark & \checkmark & &  1.076$\pm$0.095 & 0.113 & \textbf{0.794} & \textbf{4.760} & 0.191 & 0.874 & \textbf{0.960} & \textbf{0.982} \\
  \checkmark & \checkmark & \checkmark & \checkmark & \textbf{1.021}$\pm$\textbf{0.069} & \textbf{0.111} & 0.806 & 4.777 & \textbf{0.190} & \textbf{0.878} & \textbf{0.960} & \textbf{0.982} \\
\midrule
  & \checkmark & \checkmark & &  \textbf{0.968}$\pm$0.098 & 0.115 & 0.839 & 4.898 & 0.194 & 0.869 & 0.958 & 0.981 \\
  \checkmark & \checkmark & \checkmark & &  1.076$\pm$\textbf{0.095} & \textbf{0.113} & \textbf{0.794} & \textbf{4.760} & \textbf{0.191} & \textbf{0.874} & \textbf{0.960} & \textbf{0.982} \\
\midrule
   & \checkmark & \checkmark & \checkmark &  \textbf{1.013}$\pm$\textbf{0.069} & 0.112 & 0.808 & \textbf{4.751} & 0.191 & 0.877 & \textbf{0.960} & \textbf{0.982} \\
  \checkmark & \checkmark & \checkmark & \checkmark & 1.021$\pm$\textbf{0.069} & \textbf{0.111} & \textbf{0.806} & 4.777 & \textbf{0.190} & \textbf{0.878} & \textbf{0.960} & \textbf{0.982} \\
  
\bottomrule
\end{tabular}}
\end{table*}

\subsubsection{The robustness against vision degradation} We then examine the robustness of DynaDepth against illumination change and moving objects, two major cases that violate the underlying assumption of the photometric loss. We simulate the illumination change by randomly alternating image contrast within a range 0.5. The moving objects are simulated by randomly inserting three 150x150 black squares. In contrast to data augmentation, we perform the perturbation for each image independently, rather than applying the same perturbation to all images in a triplet. Results are given in Table~\ref{tb:abl_vision}. Under illumination change, the accuracy of Monodepth2 degrades as expected, while DynaDepth rescues the accuracy to a certain degree and maintains the correct absolute scales. EKF improves almost all metrics in this case, and using both EKF and $L_{vg}$ achieves the best scale and AbsRel. However, the model without $L_{vg}$ obtains the best performance on most metrics. The reason may be the dependence of $L_{vg}$ on the visual data, which is more sensitive to image qualities. When there exist moving objects, Monodepth2 fails completely. Using DynaDepth without EKF and $L_{vg}$ improves the up-to-scale accuracy a little bit, but the results are still far from expected. Using EKF significantly improves the up-to-scale results, while it is still hard to learn the scale given the difficulty of the task. In this case, using $L_{vg}$ is shown to provide strong scale supervision and achieve a good scale result.

\begin{table*}[t]
\caption{Ablation results of the robustness against vision degradation on the simulated data from KITTI. The best results are shown in \textbf{bold}. IC and MO denote the two investigated vision degradation types, i.e., illumination change and moving objects. - means item not available. $^\dagger$ denotes our reproduced results.}
\label{tb:abl_vision}
\resizebox{\textwidth}{!}{
\begin{tabular}{c|cc|cc|cccc|ccc}
\toprule
 \multirow{2}*{Methods} &\multirow{2}*{EKF} & \multirow{2}*{$L_{vg}$} & \multirow{2}*{Type}  & \multirow{2}*{Scale} & \multicolumn{4}{c|}{Error$\downarrow$} & \multicolumn{3}{c}{Accuracy$\uparrow$}\\
\cline{6-9}
\cline{10-12}
 ~ & ~ & ~ & ~ & ~ & AbsRel & SqRel & RMSE & RMSE$_{log}$ & $\sigma<1.25$ & $\sigma<1.25^2$ & $\sigma<1.25^3$ \\
 \midrule
 Monodepth2$\dagger$~\cite{godard2019digging} & - & - & IC & 27.701$\pm$0.096 & 0.127 & 0.976 & 5.019 & 0.220 & 0.855 & 0.946 & 0.972 \\
 DynaDepth &  &  & IC & 1.036$\pm$0.099 & 0.124 & \textbf{0.858} & 4.915 & 0.226 & 0.852 & 0.950 & 0.977 \\
 DynaDepth & \checkmark &  & IC & 0.946$\pm$0.089 & 0.123 & 0.925 & \textbf{4.866} & \textbf{0.196} & \textbf{0.863} & \textbf{0.957} & \textbf{0.981} \\
 DynaDepth & \checkmark & \checkmark & IC & \textbf{1.019}$\pm$\textbf{0.074} & \textbf{0.121} & 0.906 & 4.950 & 0.217 & 0.859 & 0.954 & 0.978 \\
 \midrule
 Monodepth2$\dagger$~\cite{godard2019digging} & - & - & MO & 0.291$\pm$0.176 & 0.257 & 2.493 & 8.670 & 0.398 & 0.584 & 0.801 & 0.897 \\
 DynaDepth &  &  & MO & 0.083$\pm$0.225 & 0.169 & 1.290 & 6.030 & 0.278 & 0.763 & 0.915 & 0.960 \\
 DynaDepth & \checkmark &  & MO & 0.087$\pm$0.119 & 0.126 & \textbf{0.861} & 5.312 & \textbf{0.210} & 0.840 & 0.948 & \textbf{0.979} \\
 DynaDepth & \checkmark & \checkmark & MO & \textbf{0.956}$\pm$\textbf{0.084} & \textbf{0.125} & 0.926 & \textbf{4.954} & 0.214 & \textbf{0.852} & \textbf{0.949} & 0.976 \\
\bottomrule
\end{tabular}}
\end{table*}

\subsubsection{The learnt ego-motion uncertainty} 
We illustrate the training progress of the ego-motion uncertainty in Fig.~\ref{fig:training}. We report the averaged covariance as the uncertainty measure. The learnt uncertainty exhibits a similar pattern as the depth error (AbsRel), meaning that the model becomes more certain about its predictions as the training continues. Of note is that only indirect supervision is provided, which justifies the effectiveness of our fusion framework. In addition, DynaDepth R50 achieves a lower uncertainty than R18, indicating that a larger model capacity also contributes to the prediction confidence, yet such difference can hardly be seen w.r.t. AbsRel. Table~\ref{tb:trans_uncertaity} presents another interesting observation. In KITTI, the axis-z denotes the forward direction. Since most test images correspond to driving forward, the magnitude of $t_z$ is significantly larger than $\{t_x,t_y\}$. Accordingly, DynaDepth shows a high confidence on $t_z$, while large variances are observed for $\{t_x,t_y\}$, potentially due to the difficulty to distinguish the noises from the small amount of translations along axis-x and axis-y.

\begin{table}[t]
\begin{minipage}{.62\textwidth}
    \centering
    \includegraphics[width=.95\textwidth]{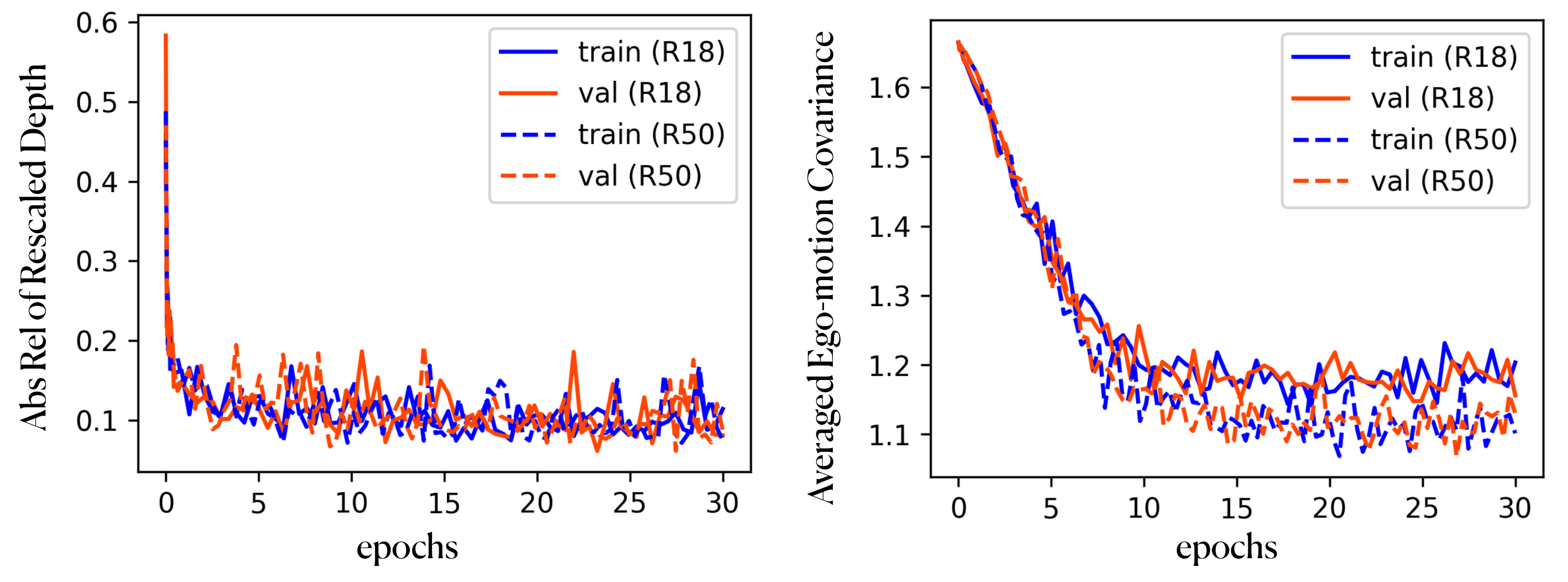}
    \captionof{figure}{The training processes w.r.t. AbsRel (left) and the averaged ego-motion covariance (right).}
    \label{fig:training}
\end{minipage}
\hfill
\parbox{.33\linewidth}{
    \centering
	\caption{The averaged magnitude $|\bar{t}|$ and the variance $\bar{\sigma}^2_{t}$ of the translation predictions along each axis.} 
	\centering
	\resizebox{.9\linewidth}{8.5mm}{
	\begin{tabular}{c c c c}
		\specialrule{0.1em}{5pt}{3pt}
        ~ & \ \ axis-x \ \  & \ \ axis-y \ \ & \ \ axis-z \ \ \\
		\midrule
		$|\bar{t}|$ & 0.017 & 0.018 & 0.811 \\
		$\bar{\sigma}^2_{t}$ & 7.559 & 5.222 & 0.105 \\
		\specialrule{0.1em}{1pt}{12pt}
	\end{tabular}}
	\label{tb:trans_uncertaity}
}
\end{table}

\section{Conclusion}
In this paper, we propose DynaDepth, a scale-aware, robust, and generalizable monocular depth estimation framework using IMU motion dynamics. Specifically, we propose an IMU photometric loss and a cross-sensor photometric consistency loss to provide dense supervision and absolution scales. In addition, we derive a camera-centric EKF framework for the sensor fusion, which also provides an ego-motion uncertainty measure under the setting of unsupervised learning. Extensive experiments support that DynaDepth is advantageous w.r.t. learning absolute scales, the generalizability, and the robustness against vision degradation.

\noindent\textbf{Acknowledgment}
This work is supported by ARC FL-170100117, DP-180103424, IC-190100031, and LE-200100049.

\clearpage
%
%
\bibliographystyle{splncs04}
\bibliography{6076}
\end{document}